\PassOptionsToPackage{main=english,russian}{babel}
\PassOptionsToPackage{T2A}{fontenc}
\documentclass[onecolumn]{vision_guided_chunk}

\usepackage{lipsum}
\usepackage{pdfpages}
\usepackage{colortbl}
\usepackage{tabulary}
\usepackage{longtable}
\usepackage{array}
\usepackage{placeins}
\usepackage{tablefootnote}
\usepackage{tcolorbox}
\usepackage{wrapfig}
\usepackage{listings}

\usepackage{breqn} 

\usepackage{pifont}
\definecolor{deeppurple}{HTML}{9e02f7}
\definecolor{forestgreen}{HTML}{2e7d43}
\usepackage{multirow}
\usepackage{tabularx}

\usepackage[utf8]{inputenc}

\usepackage{arydshln}

\usepackage{xspace} 
\usepackage{makecell} 
\usepackage{multirow}

\usepackage[framemethod=TikZ]{mdframed}
\usepackage{tcolorbox}
\usepackage{multicol}
\usepackage{dblfloatfix}

\newtcolorbox{mybox}[2][]{
  colback=white, 
  colframe=lightblue,
  fonttitle=\bfseries,
  coltitle=black,  
  title=#2, 
  #1
}

\definecolor{ayad}{RGB}{148, 156, 229} 
\definecolor{ayadsymbol}{RGB}{76, 110, 230} 
\definecolor{lightblue}{RGB}{211, 227, 252} 
\definecolor{bgblue}{RGB}{247, 250, 255} 
\newcommand*\colourcheck[1]{%
  \expandafter\newcommand\csname #1check\endcsname{\textcolor{#1}{\ding{52}}}%
}
\newcommand*\colourcross[1]{%
  \expandafter\newcommand\csname #1cross\endcsname{\textcolor{#1}{\ding{55}}}%
}
\DeclareSymbolFont{extraup}{U}{zavm}{m}{n}
\DeclareMathSymbol{\vardiamond}{\mathalpha}{extraup}{87}
\definecolor{ayadsymbol}{RGB}{76, 110, 230} 

\colourcheck{blue}
\colourcheck{green}
\colourcheck{red}

\colourcross{blue}
\colourcross{green}
\colourcross{red}
\usepackage{nicematrix}
\usepackage{comment}
\usepackage{amssymb}

\setcitestyle{number,square}

\title{The Instruction Gap: LLMs get lost in Following Instruction}

\multiauthors

\author{
    name={Vishesh Tripathi$^\circ$}
}

\author{
    name={Uday Allu$^\circ$}
}

\author{
   name={Biddwan Ahmed$^\dagger$}
}

\affiliations{
    AI Research Team, Yellow.ai}

\date{\today}
\abstract{Large Language Models (LLMs) have shown remarkable capabilities in natural language understanding and generation, yet their deployment in enterprise environments reveals a critical limitation: inconsistent adherence to custom instructions. This study presents a comprehensive evaluation of 13 leading LLMs across instruction compliance, response accuracy, and performance metrics in real-world RAG (Retrieval-Augmented Generation) scenarios. Through systematic testing with samples and enterprise-grade evaluation protocols, we demonstrate that instruction following varies dramatically across models, with Claude-Sonnet-4 and GPT-5 achieving the highest results. Our findings reveal the "instruction gap" - a fundamental challenge where models excel at general tasks but struggle with precise instruction adherence required for enterprise deployment. This work provides practical insights for organizations deploying LLM-powered solutions and establishes benchmarks for instruction-following capabilities across major model families.
}

\renewcommand{\FONTauthors}{\bfseries\Large}

\begin{document}

\footnotetext[1]{$^\circ$First authors}
\footnotetext[2]{$^\dagger$Senior authors}
\footnotetext[3]{\{vishesh.tripathi1,,uday,biddwan\} @yellow.ai}

\section{Introduction}

The rapid adoption of Large Language Models (LLMs) ~\citep{NIPS2017_3f5ee243} in enterprise environments has unveiled a critical challenge: while these models demonstrate exceptional capabilities in general language tasks, they often struggle to consistently follow specific, custom instructions required for business applications. This phenomenon, which we term the "instruction gap," represents a fundamental barrier to reliable LLM deployment in production systems.

Enterprise applications demand strict adherence ~\citep{snorkel2024enterprise} to custom guidelines, brand voice requirements, response formatting rules, and content boundaries. Unlike academic benchmarks that evaluate general capabilities, real-world deployment scenarios require models to navigate complex, multi-faceted instructions while maintaining high accuracy and consistency. Advanced evaluation frameworks and providing tailored enterprise-specific prompts, we seek to empower practitioners to assess LLMs more effectively~\citep{enterprise2024benchmarks}. Evaluating LLMs gives businesses and practitioners the insights to tune these models so they are properly calibrated to serve the AI and the specific needs of their deployments~\citep{aisera2024evaluation}

This study addresses the instruction gap through comprehensive evaluation of 13 leading LLMs across two dimensions: custom instruction compliance and response accuracy. We develop novel evaluation methodologies that address common inconsistencies in LLM assessment and provide practical insights for improving instruction adherence in production deployments.

\begin{figure}[!t]
\centering
\includegraphics[width=0.9\textwidth]{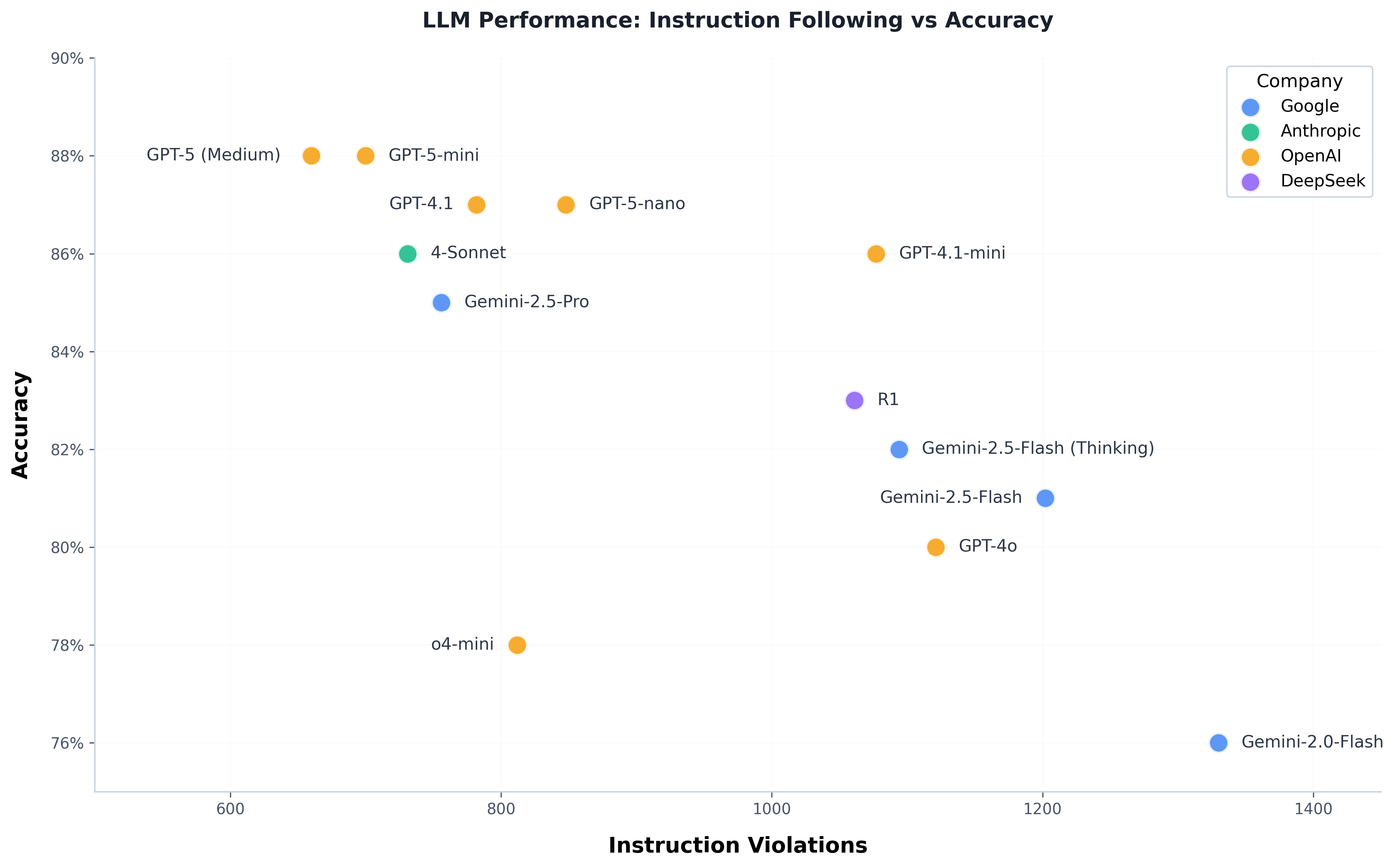}
\caption{Performance comparison of large language models on enterprise RAG scenarios across instruction following and response accuracy metrics.}
\label{fig:experimental_workflow}
\end{figure}

\subsection{Contributions}

Our key contributions include:
\begin{enumerate}
\item \textbf{Comprehensive Instruction-Following Benchmark}: We present the first systematic evaluation of instruction compliance across 13 major LLMs using real-world enterprise scenarios.
\item \textbf{Instruction Gap Analysis}: We quantify and characterize the instruction gap phenomenon, demonstrating significant variance in compliance rates across model families.
\item \textbf{Performance-Cost Trade-off Analysis}: We establish comprehensive benchmarks balancing instruction compliance, response quality, latency, and cost considerations for production deployment decisions.
\end{enumerate}

\section{Background and Related Work}

\subsection{Instruction Following in Large Language Models}

The foundation of instruction following in LLMs was established by InstructGPT~\citep{ouyang2022training}, which introduced Reinforcement Learning ~\citep{ReinforcementLearning} from Human Feedback (RLHF) ~\citep{RLHF-Deep} for aligning language models with human instructions. This seminal work demonstrated that instruction-tuned models significantly outperform their base counterparts in following user intentions, even when substantially smaller in scale.

Building on RLHF foundations, Constitutional AI~\citep{bai2022constitutional} introduced AI feedback (RLAIF) as an alternative to human feedback, enabling precise control of AI behavior through constitutional principles. This approach is particularly relevant for enterprise RAG systems where safety and compliance alignment are paramount.

Recent advances in instruction tuning include scaling studies~\citep{chung2022scaling}, which demonstrated dramatic improvements from scaling tasks, model size, and chain-of-thought data. However, these works primarily focus on general instruction following capabilities rather than enterprise-specific compliance requirements.

\subsection{RAG Systems and Enterprise Deployment}

Retrieval-Augmented Generation has emerged as a critical architecture for enterprise LLM deployment~\citep{lewis2020retrieval}. Recent advances include Active RAG~\citep{jiang2023active}, which dynamically decides when and what to retrieve during generation, and Corrective RAG~\citep{yan2024corrective}, which includes confidence scoring and web search augmentation for improved robustness.

Studies show that retrieval precision drops by up to 30\% in noisy datasets. To combat this, tools like knowledge graph augmentation and dynamic query re-weighting ensure the system focuses on contextually relevant information~\citep{chitika2024rag}.Despite its advantages, deploying RAG systems is not without its challenges. Issues related to data quality, retrieval efficiency, contextual understanding, model bias, and scalability can hinder the effectiveness~\citep{prajna2024rag}

Enterprise deployment research reveals systematic challenges including data quality management, infrastructure scaling complexity, and security concerns~\citep{enterprise2024challenges}. These studies highlight the gap between academic benchmarks and real-world deployment requirements, particularly regarding instruction adherence in production environments.

\subsection{LLM Evaluation Methodologies}

The emergence of LLM-as-a-Judge evaluation~\citep{zheng2023judging} has provided scalable alternatives to human evaluation. MT-Bench and Chatbot Arena established frameworks for systematic LLM evaluation, while identifying critical biases including position bias, verbosity bias, and self-enhancement bias.

Recent work on Prometheus 2~\citep{kim2024prometheus} represents the first open-source evaluator achieving high correlation with GPT-4 and human judgments. G-Eval~\citep{liu2023geval} introduced chain-of-thoughts evaluation with form-filling paradigms, achieving significant improvements over previous automated evaluation methods.

However, existing evaluation frameworks primarily focus on general capabilities rather than specific instruction compliance metrics required for enterprise applications.

\subsection{Instruction Gap and Compliance Research}

Systematic instruction following failures have been documented in IFEval~\citep{zhou2023instruction}, which introduced 25 types of verifiable instructions revealing failures in basic constraints like word count and keyword inclusion. LLMBar~\citep{zeng2024llmbar} demonstrated that LLM evaluators can be misled by engaging content that masks instruction violations.

Recent work on instruction hierarchy~\citep{wallace2024instruction} addresses prompt injection vulnerabilities through explicit instruction prioritization. However, enterprise-specific compliance challenges including regulatory requirements, content governance, and brand consistency remain underexplored in current literature.

\section{Experiment}

Our methodology follows a three-stage pipeline designed to assess instruction compliance in RAG scenarios. First, we configure each LLM with specific enterprise persona instructions that define behavioral guidelines, response formatting requirements, and content boundaries. Second, we present the models with user queries alongside relevant knowledge snippets, prompting them to generate responses that adhere to both the provided instructions and RAG constraints. Finally, we employ an automated evaluator to assess instruction violations and response quality, creating a systematic framework for measuring the instruction gap across different model families.

Figure~\ref{fig:experimental_workflow} illustrates our complete experimental workflow, demonstrating how we systematically test instruction following capabilities across different LLM families in enterprise RAG scenarios.

\subsection{Dataset}

We constructed a high-quality evaluation dataset comprising 600 carefully curated queries representative of real-world enterprise RAG scenarios. The dataset was derived from our internal benchmark through manual inspection and quality assurance processes to ensure representativeness and challenge diversity.

Unlike synthetic datasets commonly used in academic benchmarks, our 600 samples represent authentic enterprise queries that our engineering team spent considerable time filtering and refining. This smaller but higher-quality approach ensures each sample reflects genuine business scenarios and instruction-following challenges encountered in production deployments, providing more meaningful insights than larger synthetic datasets.The dataset design focuses on RAG summarization tasks, where instruction following becomes particularly critical due to the large context windows and complex information synthesis requirements. Each query is paired with relevant knowledge snippets and specific instructions that models must follow while generating responses.

Our instruction set encompasses five distinct enterprise personas: IT Support Agent, Teacher's Support Agent, Product Support Agent, Tech Support Agent, and Service Platform Agent. Each persona includes detailed behavioral guidelines, response formatting requirements, tone specifications, and content boundaries that reflect real-world enterprise deployment scenarios. The complete instruction examples are provided in the Appendix.

The queries span various complexity levels, from simple factual questions to complex multi-step reasoning tasks requiring synthesis of multiple information sources. This diversity ensures comprehensive evaluation of instruction adherence across different cognitive demands and contexts. Data quality was ensured through multiple validation steps: expert review of instruction clarity, verification of query-knowledge alignment, and pilot testing with human evaluators to establish ground truth expectations for instruction compliance.

\begin{figure}[!t]
\centering
\includegraphics[width=0.9\textwidth]{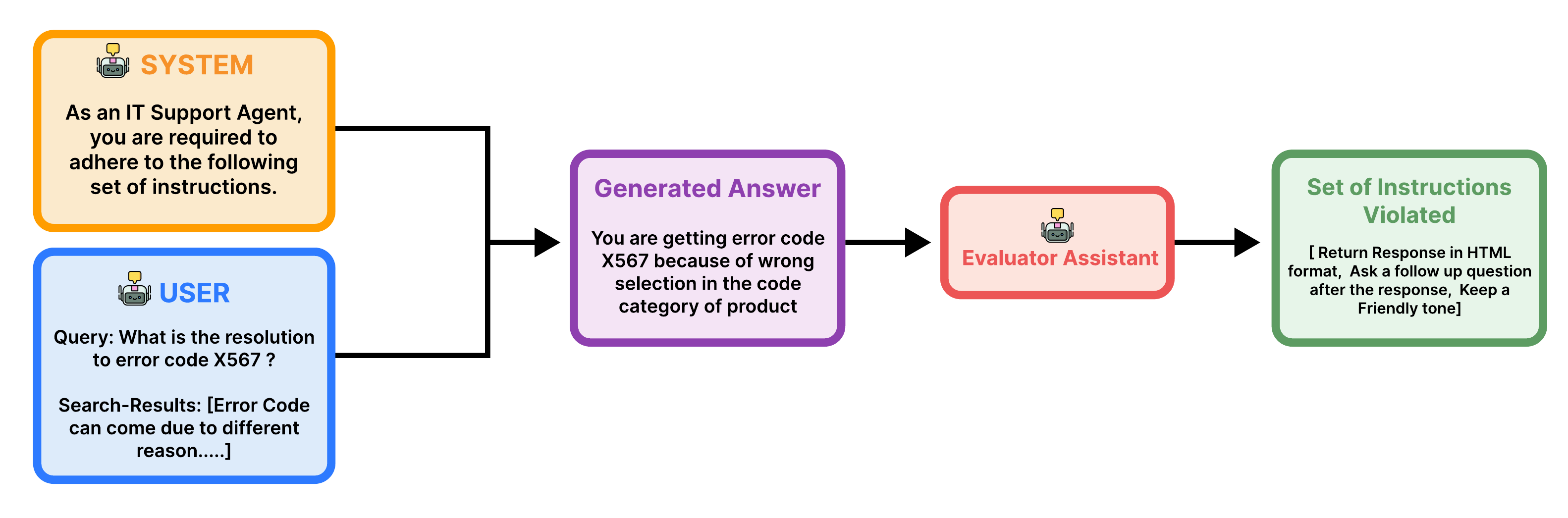}
\caption{Experimental Workflow Overview.}
\label{fig:experimental_workflow}
\end{figure}

\subsection{Evaluation Metrics}

We employ a comprehensive evaluation framework addressing both instruction compliance and response quality through automated LLM-based assessment.

\subsubsection{Instruction Violation Assessment}

Our primary metric quantifies instruction violations through systematic rule-based evaluation. We developed a specialized prompt for Evaluator model that analyzes LLM responses against the given instructions, identifying specific violations across multiple dimensions:
\begin{itemize}
\item \textbf{Content Scope Violations}: Responses outside the designated domain or expertise area
\item \textbf{Format Violations}: Deviations from specified response structure, length constraints, or formatting requirements
\item \textbf{Tone and Style Violations}: Inconsistencies with prescribed communication style, empathy requirements, or brand voice
\item \textbf{Procedural Violations}: Failures to follow specified response patterns, escalation procedures, or interaction protocols
\end{itemize}

\subsubsection{Response Quality Evaluation}

\begin{itemize}
\item \textbf{Correct}: The model provides accurate information based on available knowledge while following all instructions
\item \textbf{Hallucinate}: The model introduces information not supported by the provided knowledge base or makes unsupported inferences
\item \textbf{Abstain}: The model appropriately acknowledges limitations when lacking sufficient information to provide a complete answer
\end{itemize}

\subsubsection{LLM-as-a-Judge Validation}

We selected Claude-4-Sonnet as our primary evaluator based on preliminary validation studies comparing multiple models against human judgment on a subset of our dataset. Claude-4-Sonnet demonstrated the highest consistency and accuracy in instruction violation detection, showing strong alignment with human evaluators' assessments.

All evaluations were conducted with a temperature setting of 0.1 wherever applicable to ensure consistency and reproducibility across multiple runs. 

\subsection{Experimental Setup and Model Configuration}

We selected a total of 10 LLMs from eight model families: OpenAI (GPT-5 Family ~\citep{gpt5family}, GPT-4.1-Family, GPT-4o~\citep{hurst2024gpt} and o4-mini~\citep{openai2025o3}), Anthropic (Claude-4-Sonnet ~\citep{sonnet4}), Google's Gemini (Gemini 2.0-Flash, Gemini 2.5-Flash, Gemini 2.5-Pro)~\citep{team2025gemini}, and DeepSeek-R1~\citep{guo2025deepseek}. This selection prioritizes the evaluation of state-of-the-art models across different architectural paradigms and training methodologies.

Our model selection encompasses both open and closed weight models to provide comprehensive coverage of the current LLM landscape.We include five reasoning models ~\citep{reasoning-in-llms} (GPT-5-Family, o4-mini, DeepSeek-R1, Gemini-2.5-Pro and Gemini 2.5-Flash with thinking) to study the effect of additional test-time computation on instruction following capability. This diverse selection allows us to analyze how different approaches to model development—from traditional scaling to inference-time reasoning—impact enterprise instruction compliance.We tested Gemini 2.5-Flash in both standard mode and thinking mode, allowing direct comparison of the same underlying model with and without reasoning capabilities. In thinking mode, we allocated a thinking budget of 5000 tokens after preliminary analysis indicated this provided optimal balance between reasoning depth and computational efficiency. This token limit was determined through pilot testing that showed diminishing returns beyond 5000 tokens while maintaining reasonable inference times.

We did not enable thinking mode for Claude-4-Sonnet.This decision allows us to establish a baseline for non-reasoning models that can compete with thinking-enabled alternatives. All models were evaluated with consistent parameters: temperature set to 0.1 wherever applicable to ensure reproducible and deterministic outputs, and identical system prompts across all models to ensure fair comparison. Model versions and API endpoints are detailed in the Appendix to ensure reproducibility of our experimental setup.

\section{Results}

\begin{table}[t]
\centering
\begin{tabular}{@{}lcccc@{}}
\toprule
\multicolumn{1}{c}{Model}        & Instructions Violation              & \multicolumn{3}{c}{Accuracy}                                                                   \\ \midrule
\multicolumn{2}{l}{}                                          & \multicolumn{1}{l}{Correct} & \multicolumn{1}{l}{Hallucinate} & \multicolumn{1}{l}{Abstain} \\ \midrule
\raisebox{-0.1ex}{\includegraphics[height=1.2em]{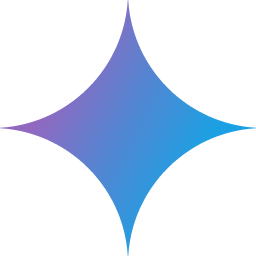}} 2.0-Flash                 & \cellcolor[HTML]{FFFFFF}1330 & \cellcolor[HTML]{FFFFFF}0.76 & \cellcolor[HTML]{FFFFFF}0.08     & \cellcolor[HTML]{FFFFFF}0.16 \\
\raisebox{-0.1ex}{\includegraphics[height=1.2em]{icons/gemini.png}} 2.5-Flash   & \cellcolor[HTML]{FFFFFF}1202 & \cellcolor[HTML]{FFFFFF}0.81 & \cellcolor[HTML]{FFFFFF}0.08     & \cellcolor[HTML]{FFFFFF}0.11 \\
\raisebox{-0.1ex}{\includegraphics[height=1.2em]{icons/gemini.png}} 2.5-Flash (Thinking) & \cellcolor[HTML]{FFFFFF}1094 & \cellcolor[HTML]{FFFFFF}0.82 & \cellcolor[HTML]{FFFFFF}0.03     & \cellcolor[HTML]{FFFFFF}0.15 \\
\raisebox{-0.1ex}{\includegraphics[height=1.2em]{icons/gemini.png}} 2.5-Pro                   & \cellcolor[HTML]{FFFFFF}756  & \cellcolor[HTML]{FFFFFF}0.85 & \cellcolor[HTML]{FFFFFF}0.04     & \cellcolor[HTML]{FFFFFF}0.11 \\
\raisebox{-0.1ex}{\includegraphics[height=1.2em]{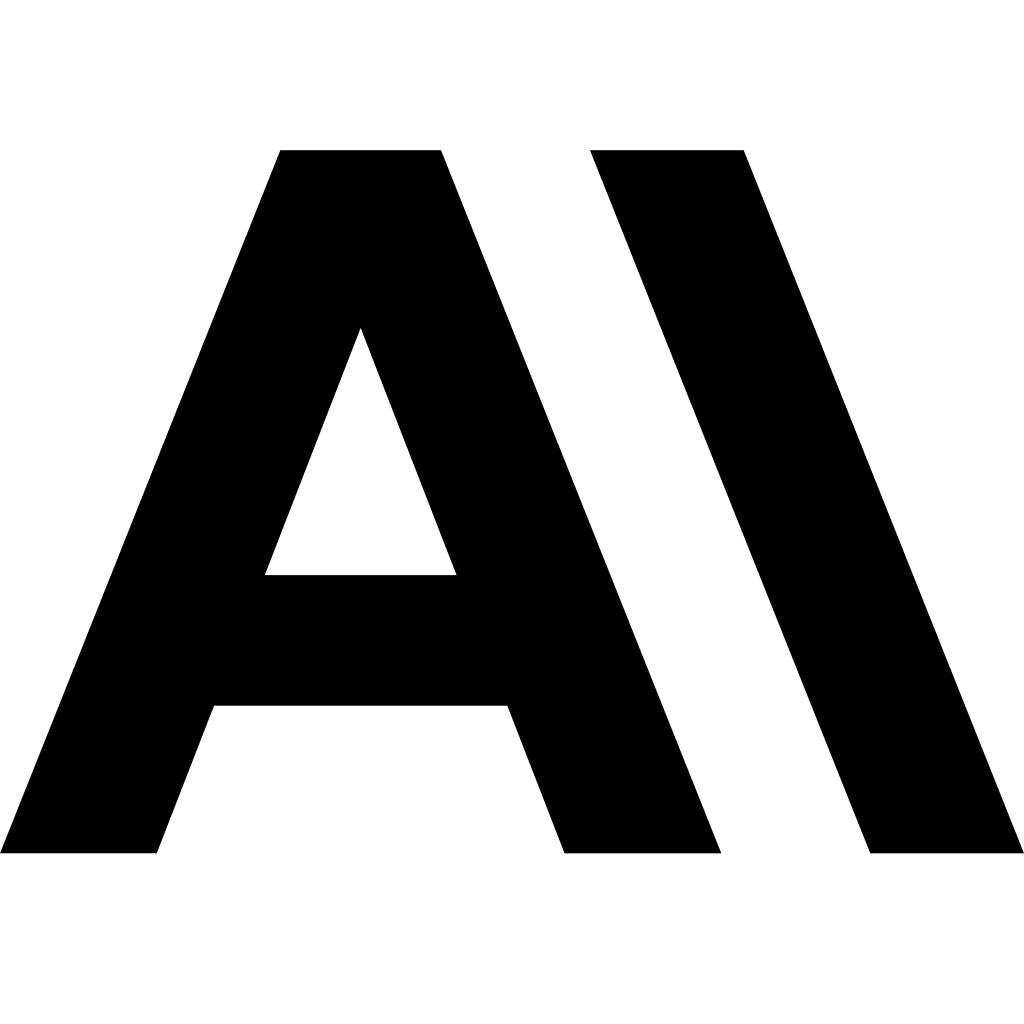}} 4-Sonnet                  & \cellcolor[HTML]{FFFFFF}731  & \cellcolor[HTML]{FFFFFF}0.86 & \cellcolor[HTML]{FFFFFF}0.04     & \cellcolor[HTML]{FFFFFF}0.09 \\
\raisebox{-0.1ex}{\includegraphics[height=1.2em]{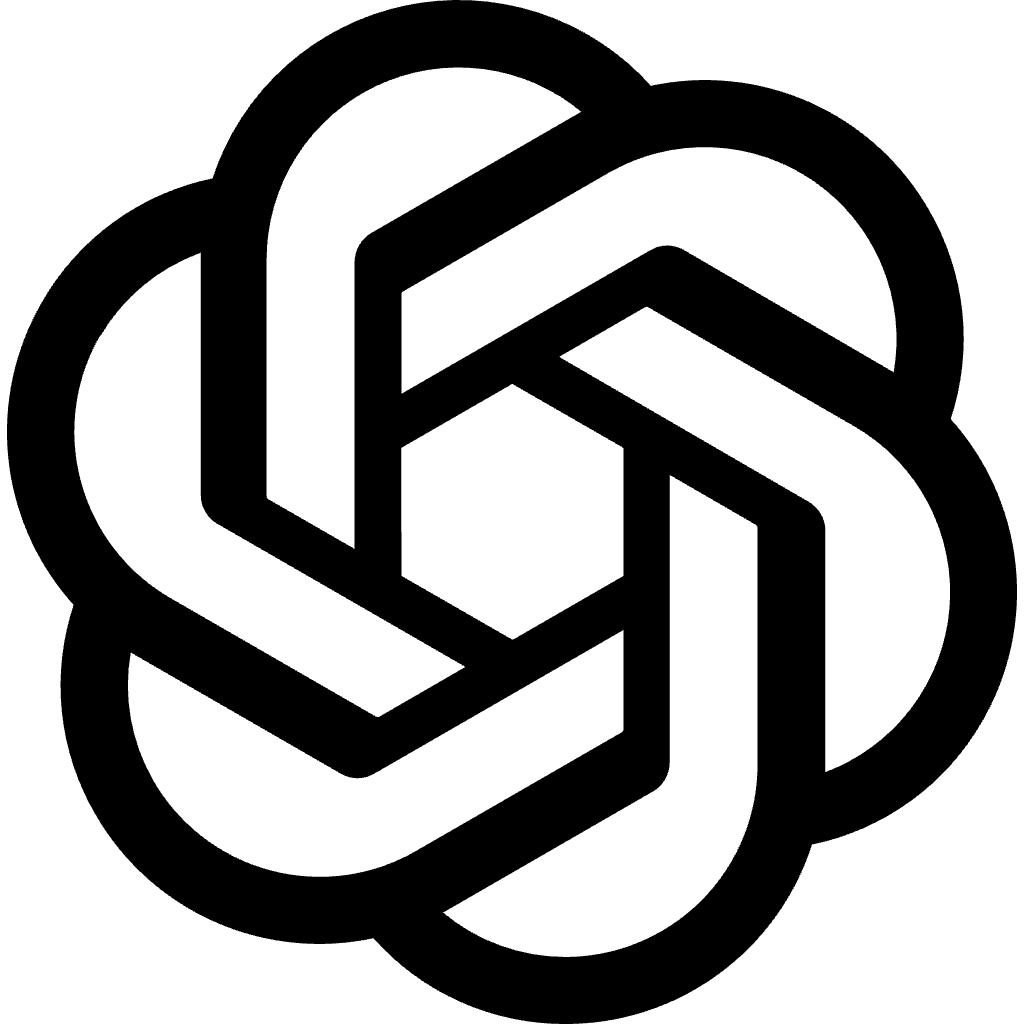}} o4-mini                          & \cellcolor[HTML]{FFFFFF}812  & \cellcolor[HTML]{FFFFFF}0.78 & \cellcolor[HTML]{FFFFFF}0.05     & \cellcolor[HTML]{FFFFFF}0.16 \\
\raisebox{-0.1ex}{\includegraphics[height=1.2em]{icons/openai.png}} 4.1                          & \cellcolor[HTML]{FFFFFF}782  & \cellcolor[HTML]{FFFFFF}0.87 & \cellcolor[HTML]{FFFFFF}0.07     & \cellcolor[HTML]{FFFFFF}0.05 \\
\raisebox{-0.1ex}{\includegraphics[height=1.2em]{icons/openai.png}} 4.1-mini                     & \cellcolor[HTML]{FFFFFF}1077 & \cellcolor[HTML]{FFFFFF}0.86 & \cellcolor[HTML]{FFFFFF}0.08     & \cellcolor[HTML]{FFFFFF}0.05 \\
\raisebox{-0.1ex}{\includegraphics[height=1.2em]{icons/openai.png}} 4o                           & \cellcolor[HTML]{FFFFFF}1121 & \cellcolor[HTML]{FFFFFF}0.80 & \cellcolor[HTML]{FFFFFF}0.05     & \cellcolor[HTML]{FFFFFF}0.14 \\
\raisebox{-0.1ex}{\includegraphics[height=1.2em]{icons/openai.png}} 5 (Medium)        & \cellcolor[HTML]{FFFFFF}660  & \cellcolor[HTML]{FFFFFF}0.88 & \cellcolor[HTML]{FFFFFF}0.03    & \cellcolor[HTML]{FFFFFF}0.09 \\
\raisebox{-0.1ex}{\includegraphics[height=1.2em]{icons/openai.png}} 5-mini                       & \cellcolor[HTML]{FFFFFF}700  & \cellcolor[HTML]{FFFFFF}0.88 & \cellcolor[HTML]{FFFFFF}0.04   & \cellcolor[HTML]{FFFFFF}0.08 \\
\raisebox{-0.1ex}{\includegraphics[height=1.2em]{icons/openai.png}} 5-nano                       & \cellcolor[HTML]{FFFFFF}848  & \cellcolor[HTML]{FFFFFF}0.87 & \cellcolor[HTML]{FFFFFF}0.06   & \cellcolor[HTML]{FFFFFF}0.07 \\
\raisebox{-0.1ex}{\includegraphics[height=1.2em]{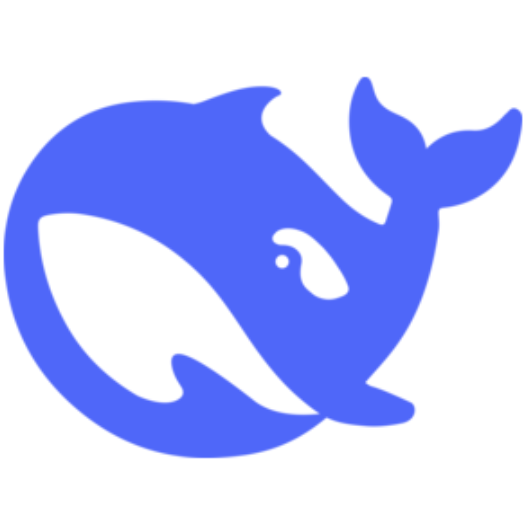}} R1                      & \cellcolor[HTML]{FFFFFF}1061 & \cellcolor[HTML]{FFFFFF}0.83 & \cellcolor[HTML]{FFFFFF}0.09     & \cellcolor[HTML]{FFFFFF}0.07 \\ \bottomrule
\end{tabular}
\caption{Instruction following and response accuracy evaluation across 13 LLMs on enterprise RAG scenarios. The table shows instruction violation counts (lower is better) and response accuracy broken down into three categories: Correct, Hallucinate, and Abstain.}
\label{tab:result_table}
\end{table}

\subsection{Instruction Violations}

Table~\ref{tab:result_table} summarizes results from our experiments. The instruction violation analysis reveals substantial differences in compliance capabilities across model families, with violation counts ranging from 660 to 1330 across our evaluation set.

GPT-5 (Medium) emerges as the top performer with only 660 violations, representing a significant advancement in instruction-following capabilities. This is followed closely by GPT-5-mini (700 violations) and Claude 4-Sonnet (731 violations). Claude 4-Sonnet remains the best-performing non-reasoning model, demonstrating superior instruction adherence without additional test-time computation.

The GPT-5 family shows remarkable consistency across model sizes, with GPT-5-nano achieving 848 violations—still competitive despite being the smallest variant. In contrast, earlier generation models show much higher violation rates, with Gemini 2.0-Flash and GPT-4o accumulating 1330 and 1121 violations respectively—nearly double the best-performing models

Notably, GPT-5 (Medium), as a reasoning model, outperforms all other reasoning models including DeepSeek R1 (1061), o4-mini (812), and Gemini-2.5-Pro (756), establishing a new benchmark for reasoning-enhanced instruction following. While previous work ~\citep{illusionOfThinking} has primarily examined the general capabilities of thinking models, our experimental results demonstrate that reasoning capabilities significantly enhance instruction-following performance. Specifically, we observe that thinking-enabled models, including GPT-5, Gemini-2.5-Pro, DeepSeek-R1, and Gemini-2.5-Flash (Thinking), consistently outperform the non-thinking models. The thinking-enabled Gemini 2.5-Flash variant shows improved compliance (1094) compared to its non-thinking counterpart (1202), indicating that reasoning processes can enhance instruction following, though the improvements are modest compared to the GPT-5 family's performance.

Our results demonstrate that reasoning capabilities significantly enhance instruction-following performance across multiple model families. However, the exceptional performance of GPT-5 models, particularly GPT-5 (Medium), suggests that advanced reasoning capabilities combined with improved training methodologies can achieve substantial improvements in instruction compliance. The strong performance of Claude-4-Sonnet and GPT-4.1 without reasoning capabilities provides important evidence that fundamental architectural optimizations remain critical for instruction following.

\subsection{Response Accuracy}

Response accuracy evaluation reveals critical patterns that diverge from instruction compliance trends. While reasoning models generally demonstrate improved instruction following, the same benefits are not consistently reflected in response accuracy for RAG tasks, with one notable exception. GPT-5 models demonstrate exceptional performance in both instruction following and response accuracy. GPT-5 (Medium) and GPT-5-mini both achieve 0.88 accuracy, the highest in our evaluation, while GPT-5-nano maintains competitive performance at 0.87. This represents a significant advancement where reasoning capabilities translate directly into improved factual accuracy.

GPT-4.1 showed strong performance (0.87) and maintains competitive accuracy despite lower instruction compliance. We observed that o4-mini, despite being a reasoning model, performs notably worse (0.78) than other models in our evaluation. Upon investigation, we found that this model tends to overthink, leading to abstention on many tasks where it could have answered easily (16\% abstain rate). This over-cautious behavior manifests as excessive uncertainty, preventing the model from providing useful responses even when sufficient information is available.

Claude 4-Sonnet, while being the best non-reasoning model for instruction following, also demonstrates strong accuracy (0.86) with low hallucination (0.04) and abstain rates (0.09), making it an excellent choice for applications where reasoning overhead is not desired. Thinking-enabled models show mixed results in accuracy metrics. Gemini-2.5-Flash with thinking demonstrates improvement over its non-thinking counterpart (0.82 vs 0.81), with abstain rate increasing by 4\% and hallucination rate decreasing by 1\%. However, these improvements are modest compared to the substantial gains observed in the GPT-5 family.

The GPT-5 family represents a breakthrough in bridging the instruction-accuracy gap, achieving both superior instruction compliance and high factual accuracy. This finding challenges our previous observation that instruction following and accuracy represent largely independent capabilities, suggesting that sufficiently advanced reasoning models can optimize for both objectives simultaneously.

The pattern of abstention rates reveals important behavioral differences across models. Some models adopt conservative strategies, preferring to acknowledge limitations rather than risk providing incorrect information. Others demonstrate more confident response patterns, providing definitive answers even in ambiguous scenarios. This trade-off between decisiveness and caution represents a fundamental consideration for enterprise deployment, where both excessive abstention and confident misinformation carry business risks. This finding shows a crucial insight: models that follow all instructions will not necessarily provide accurate answers, and conversely, models with high accuracy may struggle with instruction compliance. The instruction gap extends beyond simple rule-following to encompass the complex relationship between procedural adherence and knowledge synthesis.

\section{Discussion}

Our findings reveal systematic patterns in instruction following failures that extend beyond simple rule compliance to fundamental architectural limitations. The significant variation in violation counts—from 660 to 1330 across models—suggests that instruction adherence reflects deeper issues in how LLMs process and prioritize multi-faceted requirements. The emergence of GPT-5 (Medium) as the top performer with only 660 violations represents a paradigm shift in instruction-following capabilities. This model demonstrates that advanced reasoning architectures can achieve substantial improvements in both instruction compliance and response accuracy, challenging the traditional trade-off between these metrics. The consistent performance across the GPT-5 family (GPT-5-mini: 700, GPT-5-nano: 848) suggests systematic architectural improvements rather than isolated optimizations.

Claude 4-Sonnet's position as the best non-reasoning model (731 violations, 0.86 accuracy) highlights the continued importance of fundamental architectural design. Its superior performance without additional test-time computation demonstrates that effective instruction following can be achieved through training methodology and model architecture optimization, providing a cost-effective alternative to reasoning-enhanced models. Our results align with documented lost in the middle behaviors where LLMs struggle to utilize information from the middle portions of long contexts ~\citep{liu2023lost}. In enterprise RAG scenarios, instructions often compete for attention with lengthy knowledge snippets, potentially causing models to lose focus on critical compliance requirements. This phenomenon manifests as a disconnect between information retrieval and utilization—models may encode instruction information but fail to leverage it effectively during generation~\citep{lu2024insights}.

The GPT-5 family's success suggests that advanced reasoning capabilities can overcome attention allocation challenges that plague earlier model generations. The substantial gap between GPT-5 (Medium) and earlier reasoning models like o4-mini (812 violations) and DeepSeek R1 (1061 violations) indicates that reasoning quality, not just reasoning presence, critically impacts instruction following performance.

Research has identified that LLMs exhibit three primary failure modes at longer context lengths repeated content generation, random irrelevant responses, and failures to follow instructions~\citep{databricks2024long}. Our observations of simple instruction violations—such as GPT-4o and Gemini 2.0-Flash failing to produce HTML-formatted responses despite explicit formatting requirements—suggest that attention mechanisms become diluted when processing complex, multi-component instructions alongside substantial context~\citep{peysakhovich2023attention}.

The mixed performance of thinking-enabled models (Gemini 2.5-Flash with thinking showing modest improvements while o4-mini demonstrating over-cautious behavior) indicates that additional computational steps alone cannot overcome fundamental attention allocation challenges. This suggests that the quality of reasoning implementation, rather than mere presence of reasoning capabilities, determines effectiveness in instruction following tasks. The superior performance of GPT-5 models may reflect architectural optimizations specifically designed for instruction prioritization and context management, while Claude 4-Sonnet's success demonstrates that these improvements can also be achieved through non-reasoning approaches. This duality provides organizations with multiple pathways for achieving reliable instruction compliance based on their specific computational and cost constraints.

\section{Conclusion}

This study presents the first comprehensive evaluation of instruction following capabilities across 10 leading LLMs in enterprise RAG scenarios. Our findings reveal significant variations in instruction compliance, with violation counts ranging from 660 to 1330, and accuracy rates spanning 0.76 to 0.88. The instruction gap phenomenon is clearly evident: models that excel in general benchmarks may struggle with specific enterprise instruction requirements. Our work establishes practical benchmarks for enterprise LLM selection and highlights critical areas for continued research. As organizations increasingly depend on LLM-powered systems, understanding and addressing instruction following limitations becomes essential for successful deployment. The enterprise AI community would benefit from continued research into specialized training approaches, adaptive evaluation frameworks, and cost-effective solutions that maintain high instruction compliance standards required for production systems.

\section{Limitations}

This study focuses specifically on RAG summarization tasks within enterprise contexts, which may not fully represent the breadth of instruction following challenges across all LLM applications. Our evaluation relies on LLM-as-a-Judge methodology, which, despite validation against human judgment, may introduce systematic biases not captured in our assessment. The five enterprise personas, while representative, do not encompass all possible business contexts where instruction following is critical.

Our experimental design was intentionally limited to zero-shot evaluation without providing examples or demonstrations to models, as we aimed to test real-world enterprise prompts given by clients for their agents and personas. Additionally, we maintained identical prompts across all models to ensure fair comparison, despite each model provider offering specific prompting guidelines and optimization strategies ~\citep{GooglePromptGuide, OpenAIPromptGuide, AnthropicPromptGuide}. While this approach reflects authentic deployment scenarios where LLMs must interpret instructions without prior examples and ensures experimental consistency, it does not capture how instruction following performance might vary under few-shot conditions, model-specific prompt optimization, or when models are provided with explicit examples of desired behavior. The absence of in-context learning scenarios and model-tailored prompting may have contributed to some of the observed instruction violations, as models might perform differently when given concrete demonstrations of the expected output format or prompts optimized for their specific architectures, though this would deviate from typical enterprise usage patterns where standardized prompts are often deployed across multiple models.

Despite identifying clear patterns in instruction violations, we were unable to isolate a single concrete mechanism explaining why LLMs consistently fail to follow specific instructions. The interplay between attention allocation, context length effects, and instruction complexity requires further investigation through mechanistic interpretability studies ~\citep{Interpretability}. Future work should explore how different prompting strategies, including few-shot demonstrations and chain-of-thought reasoning, might mitigate the instruction following failures observed in this study.

\bibliography{addon}

@article{ouyang2022training,
  title={Training language models to follow instructions with human feedback},
  author={Ouyang, Long and Wu, Jeff and Jiang, Xu and Almeida, Diogo and Wainwright, Carroll L and Mishkin, Pamela and Zhang, Chong and Agarwal, Sandhini and Slama, Katarina and Ray, Alex and others},
  journal={Advances in Neural Information Processing Systems},
  volume={35},
  pages={27730--27744},
  year={2022}
}

@article{bai2022constitutional,
  title={Constitutional ai: Harmlessness from ai feedback},
  author={Bai, Yuntao and Kadavath, Saurav and Kundu, Sandipan and Askell, Amanda and Kernion, Jackson and Jones, Andy and Chen, Anna and Goldie, Anna and Mirhoseini, Azalia and McKinnon, Cameron and others},
  journal={arXiv preprint arXiv:2212.08073},
  year={2022}
}

@article{chung2022scaling,
  title={Scaling instruction-finetuned language models},
  author={Chung, Hyung Won and Hou, Le and Longpre, Shayne and Zoph, Barret and Tay, Yi and Fedus, William and Li, Yunxuan and Wang, Xuezhi and Dehghani, Mostafa and Brahma, Siddhartha and others},
  journal={arXiv preprint arXiv:2210.11416},
  year={2022}
}

@inproceedings{lewis2020retrieval,
  title={Retrieval-augmented generation for knowledge-intensive nlp tasks},
  author={Lewis, Patrick and Perez, Ethan and Piktus, Aleksandra and Petroni, Fabio and Karpukhin, Vladimir and Goyal, Naman and K{\"u}ttler, Heinrich and Lewis, Mike and Yih, Wen-tau and Rockt{\"a}schel, Tim and others},
  booktitle={Advances in neural information processing systems},
  volume={33},
  pages={9459--9474},
  year={2020}
}

@article{jiang2023active,
  title={Active retrieval augmented generation},
  author={Jiang, Zhengbao and Xu, Frank F and Gao, Luyu and Sun, Zhiqing and Liu, Qian and Dwivedi-Yu, Jane and Yang, Yiming and Callan, Jamie and Neubig, Graham},
  journal={arXiv preprint arXiv:2305.06983},
  year={2023}
}

@article{yan2024corrective,
  title={Corrective retrieval augmented generation},
  author={Yan, Shi-Qi and Gu, Jia-Chen and Zhu, Yun and Ling, Zhi-Hao},
  journal={arXiv preprint arXiv:2401.15884},
  year={2024}
}

@inproceedings{zheng2023judging,
  title={Judging LLM-as-a-Judge with MT-Bench and Chatbot Arena},
  author={Zheng, Lianmin and Chiang, Wei-Lin and Sheng, Ying and Zhuang, Siyuan and Wu, Zhanghao and Zhuang, Yonghao and Lin, Zi and Li, Zhuohan and Li, Dacheng and Xing, Eric and others},
  booktitle={Thirty-seventh Conference on Neural Information Processing Systems Datasets and Benchmarks Track},
  year={2023}
}

@article{kim2024prometheus,
  title={Prometheus 2: An Open Source Language Model Specialized in Evaluating Other Language Models},
  author={Kim, Seungone and Suk, Juyoung and Cho, Ji Yong and Jang, Joel and Longpre, Shayne and Lee, Hong-in and Yun, Sung Ju and Shin, Hwaran and Kang, Sangdoo and Kim, Seongjin and others},
  journal={arXiv preprint arXiv:2405.01535},
  year={2024}
}

@inproceedings{liu2023geval,
  title={G-Eval: NLG Evaluation using GPT-4 with Better Human Alignment},
  author={Liu, Yang and Iter, Dan and Xu, Yichong and Wang, Shuohang and Xu, Ruochen and Zhu, Chenguang},
  booktitle={Proceedings of the 2023 Conference on Empirical Methods in Natural Language Processing},
  pages={2511--2522},
  year={2023}
}

@article{zhou2023instruction,
  title={Instruction-Following Evaluation for Large Language Models},
  author={Zhou, Jeffrey and Lu, Tianjian and Mishra, Swaroop and Brahma, Siddhartha and Basu, Sujoy and Luan, Yi and Zhou, Denny and Hou, Le},
  journal={arXiv preprint arXiv:2311.07911},
  year={2023}
}

@inproceedings{zeng2024llmbar,
  title={LLMBar: Evaluating Large Language Models at Evaluating Instruction Following},
  author={Zeng, Zhiyuan and Yu, Jiatong and Gao, Tianyu and Yu, Yang and Zhang, Yue and He, Danqi},
  booktitle={The Twelfth International Conference on Learning Representations},
  year={2024}
}

@article{wallace2024instruction,
  title={The Instruction Hierarchy: Training LLMs to Prioritize Privileged Instructions},
  author={Wallace, Eric and Zhao, Kai and Ngo, Chien-ping and Ellis, Kevin},
  journal={arXiv preprint arXiv:2404.13208},
  year={2024}
}

@misc{enterprise2024challenges,
  title={Enterprise LLM Deployment: Challenges and Solutions},
  author={{AI21 Labs and Coralogix}},
  howpublished={\url{https://www.ai21.com/blog/enterprise-generative-ai-key-challenges-and-how-to-solve-them}},
  year={2024},
  note={Accessed: 2025-01-01}
}

@misc{illusionOfThinking,
      title={The Illusion of Thinking: Understanding the Strengths and Limitations of Reasoning Models via the Lens of Problem Complexity}, 
      author={Parshin Shojaee and Iman Mirzadeh and Keivan Alizadeh and Maxwell Horton and Samy Bengio and Mehrdad Farajtabar},
      year={2025},
      eprint={2506.06941},
      archivePrefix={arXiv},
      primaryClass={cs.AI},
      url={https://arxiv.org/abs/2506.06941}, 
}

@misc{openai2025o3,
	author = {OpenAI},
    year={2025},
	title = {{O}pen{A}{I} o3 and o4-mini {S}ystem {C}ard --- openai.com},
	howpublished = {\url{https://openai.com/index/o3-o4-mini-system-card/}},
	note = {[Accessed 08-05-2025]},
}

@article{hurst2024gpt,
  title={Gpt-4o system card},
  author={Hurst, Aaron and Lerer, Adam and Goucher, Adam P and Perelman, Adam and Ramesh, Aditya and Clark, Aidan and Ostrow, AJ and Welihinda, Akila and Hayes, Alan and Radford, Alec and others},
  journal={arXiv preprint arXiv:2410.21276},
  year={2024}
}

@article{gpt5family,
  title={GPT-5 Family},
  author={OpenAI},
  journal={https://openai.com/index/introducing-gpt-5/},
  year={2025}
}

@misc{team2025gemini,
  title={Gemini 2.5: Pushing the Frontier with Advanced Reasoning , Multimodality, Long Context, and Next Generation Agentic Capabilities},
  author={Google},
  howpublished={https://storage.googleapis.com/deepmind-media/gemini/gemini_v2_5_report.pdf},
  year={2025}
}

@misc{GooglePromptGuide,
  title={Gemini Prompt Guide},
  author={Google},
  howpublished={https://ai.google.dev/gemini-api/docs/prompting-strategies},
  year={2025}
}

@misc{OpenAIPromptGuide,
  title={OpenAI Prompt Guide},
  author={OpenAI},
  howpublished={https://help.openai.com/en/articles/6654000-best-practices-for-prompt-engineering-with-the-openai-api},
  year={2025}
}

@misc{AnthropicPromptGuide,
  title={Anthropic Prompt Guide},
  author={Anthropic},
  howpublished={https://docs.anthropic.com/en/docs/build-with-claude/prompt-engineering/claude-4-best-practices},
  year={2025}
}

@misc{Interpretability,
  title={The Urgency of Interpretability},
  author={Anthropic},
  howpublished={https://www.darioamodei.com/post/the-urgency-of-interpretability},
  year={2025}
}

@ARTICLE{ReinforcementLearning,
  author={Sutton, R.S. and Barto, A.G.},
  journal={IEEE Transactions on Neural Networks}, 
  title={Reinforcement Learning: An Introduction}, 
  year={1998},
  volume={9},
  number={5},
  pages={1054-1054},
  doi={10.1109/TNN.1998.712192}}

@inproceedings{RLHF-Deep,
 author = {Christiano, Paul F and Leike, Jan and Brown, Tom and Martic, Miljan and Legg, Shane and Amodei, Dario},
 booktitle = {Advances in Neural Information Processing Systems},
 editor = {I. Guyon and U. Von Luxburg and S. Bengio and H. Wallach and R. Fergus and S. Vishwanathan and R. Garnett},
 pages = {},
 publisher = {Curran Associates, Inc.},
 title = {Deep Reinforcement Learning from Human Preferences},
 url = {https://proceedings.neurips.cc/paper_files/paper/2017/file/d5e2c0adad503c91f91df240d0cd4e49-Paper.pdf},
 volume = {30},
 year = {2017}
}

@inproceedings{reasoning-in-llms,
author = {Wei, Jason and Wang, Xuezhi and Schuurmans, Dale and Bosma, Maarten and Ichter, Brian and Xia, Fei and Chi, Ed H. and Le, Quoc V. and Zhou, Denny},
title = {Chain-of-thought prompting elicits reasoning in large language models},
year = {2022},
isbn = {9781713871088},
publisher = {Curran Associates Inc.},
address = {Red Hook, NY, USA},
booktitle = {Proceedings of the 36th International Conference on Neural Information Processing Systems},
articleno = {1800},
numpages = {14},
location = {New Orleans, LA, USA},
series = {NIPS '22}
}

@inproceedings{NIPS2017_3f5ee243,
 author = {Vaswani, Ashish and Shazeer, Noam and Parmar, Niki and Uszkoreit, Jakob and Jones, Llion and Gomez, Aidan N and Kaiser, \L ukasz and Polosukhin, Illia},
 booktitle = {Advances in Neural Information Processing Systems},
 editor = {I. Guyon and U. Von Luxburg and S. Bengio and H. Wallach and R. Fergus and S. Vishwanathan and R. Garnett},
 pages = {},
 publisher = {Curran Associates, Inc.},
 title = {Attention is All you Need},
 url = {https://proceedings.neurips.cc/paper_files/paper/2017/file/3f5ee243547dee91fbd053c1c4a845aa-Paper.pdf},
 volume = {30},
 year = {2017}
}

@misc{sonnet4,
  title={Claude-4},
  author={Anthropic},
  howpublished={https://www.anthropic.com/news/claude-4},
  year={2025}
}

@article{guo2025deepseek,
  title={Deepseek-r1: Incentivizing reasoning capability in llms via reinforcement learning},
  author={Guo, Daya and Yang, Dejian and Zhang, Haowei and Song, Junxiao and Zhang, Ruoyu and Xu, Runxin and Zhu, Qihao and Ma, Shirong and Wang, Peiyi and Bi, Xiao and others},
  journal={arXiv preprint arXiv:2501.12948},
  year={2025}
}

@article{liu2023lost,
  title={Lost in the Middle: How Language Models Use Long Contexts},
  author={Liu, Nelson F and Lin, Kevin and Hewitt, John and Paranjape, Ashwin and Belinkov, Yonatan and Liang, Percy},
  journal={arXiv preprint arXiv:2307.03172},
  year={2023}
}

@article{lu2024insights,
  title={Insights into LLM Long-Context Failures: When Transformers Know but Don't Tell},
  author={Lu, Taiming and Gao, Muhan and Yu, Kuai and Byerly, Adam and Khashabi, Daniel},
  journal={arXiv preprint arXiv:2406.14673},
  year={2024}
}

@article{databricks2024long,
  title={Long Context RAG Performance of LLMs},
  author={Databricks},
  journal={Databricks Blog},
  year={2024},
  url={https://www.databricks.com/blog/long-context-rag-performance-llms}
}

@article{peysakhovich2023attention,
  title={Attention sorting combats recency bias in long context language models},
  author={Peysakhovich, Alexander and Lerer, Adam},
  journal={arXiv preprint arXiv:2310.01427},
  year={2023}
}

@misc{snorkel2024enterprise,
  title={LLM evaluation in enterprise applications: a new era in ML},
  author={Snorkel AI},
  year={2024},
  url={https://snorkel.ai/llm-evaluation/},
  note={Accessed: 2025-01-07}
}

@article{enterprise2024benchmarks,
  title={Enterprise Benchmarks for Large Language Model Evaluation},
  author={IBM Research and MIT-IBM Watson AI Lab},
  journal={arXiv preprint arXiv:2410.12857},
  year={2024},
  url={https://arxiv.org/html/2410.12857v1}
}

@misc{aisera2024evaluation,
  title={LLM Evaluation: Key Metrics, Best Practices and Frameworks},
  author={Aisera},
  year={2024},
  url={https://aisera.com/blog/llm-evaluation/},
  note={Accessed: 2025-01-07}
}

@misc{chitika2024rag,
  title={Retrieval-Augmented Generation: Challenges \& Solutions},
  author={Chitika},
  year={2025},
  url={https://www.chitika.com/rag-challenges-and-solution/},
  note={Accessed: 2025-01-07}
}

@misc{prajna2024rag,
  title={Overcoming Challenges in RAG Deployment: A Comprehensive Guide},
  author={PrajnaAI},
  journal={Medium},
  year={2024},
  url={https://prajnaaiwisdom.medium.com/overcoming-challenges-in-rag-deployment-a-comprehensive-guide-61e1fe937ce4},
  note={Accessed: 2025-01-07}
}

\clearpage
\newpage
\appendix

\section{Appendix}

\subsection{Instructions Used Examples \\} 

\begin{mdframed}[
    frametitle={IT Support Agent},
    frametitlebackgroundcolor=lightblue!50,
    backgroundcolor=lightblue!10,
    linecolor=lightblue!50,
    linewidth=1pt,
    roundcorner=3pt,
    innertopmargin=5pt,
    innerbottommargin=5pt,
    innerleftmargin=5pt,
    innerrightmargin=5pt
]

{\small
{\itshape You are an Aperica IT support agent. Help users only with Aperica-related IT issues. Follow these rules exactly:

Scope: If a request is outside Aperica IT support (e.g., billing, cooking, general advice), reply “Sorry, I can’t help you with that.”.\\
Directness: Answer the user’s exact question without assumption.\\
Unable to answer: If you can’t provide a specific, accurate Aperica IT answer, reply “This might need help from a specialist, as I’m unable to provide the specific information you’re looking for right now.”\\
Clarity and length: Use simple, conversational language. Limit responses to four sentences and no more than 100 words.\\
Avoid jargon: Do not use acronyms or internal terms without a brief explanation.\\
Response structure must always follow this pattern:\\
Empathetic acknowledgment\\
Brief direct answer\\
Optional link suggestion (if applicable) \\
Thoughtful follow-up question \\
}}
\end{mdframed}

\begin{mdframed}[
    frametitle={Teacher's Support Agent},
    frametitlebackgroundcolor=lightblue!50,
    backgroundcolor=lightblue!10,
    linecolor=lightblue!50,
    linewidth=1pt,
    roundcorner=3pt,
    innertopmargin=5pt,
    innerbottommargin=5pt,
    innerleftmargin=5pt,
    innerrightmargin=5pt
]

{\small
{\itshape As an AI assistant on the NEI Support website for National Education Institute, respond strictly based on the user’s preselected:

Platform: NEI, NEW, or MyNEI\\
Role: Teacher, Student, or Curriculum Admin\\
Program: e.g., Literacy, Science, Math

Your responses must:\\
Be concise and actionable. Limit each step to one clear instruction.\\
Use numbered lists for multi-step guidance.\\
Format with single line breaks between steps—no asterisks, hashtags, or bullet symbols.\\
Employ direct, action-oriented phrasing, starting each step with a verb (e.g., “Select,” “Click,” “Enter”).\\
For known error codes (e.g., 1213, 5009): \\
1. Provide a numbered troubleshooting procedure. \\ 
2. If unresolved or requiring higher permissions, add: “If this does not resolve the issue, contact [Platform] Support or your Curriculum Admin.” \\
Tailor solutions to the user’s role: include only steps the user can perform. \\
For tasks needing admin access (system settings, user management), end with: “Please contact your Curriculum Admin to complete this action.” 
}}
\end{mdframed}

\begin{mdframed}[
    frametitle={Product Support Agent},
    frametitlebackgroundcolor=lightblue!50,
    backgroundcolor=lightblue!10,
    linecolor=lightblue!50,
    linewidth=1pt,
    roundcorner=3pt,
    innertopmargin=5pt,
    innerbottommargin=5pt,
    innerleftmargin=5pt,
    innerrightmargin=5pt
]

{\small
{\itshape The following is a conversation with a goal-oriented, friendly, supportive chatbot for Zesto Care. The chatbot assists new parents with technical support and product or subscription questions about Owlet Care services. Follow these guidelines:

Tone and Style\\
Always use a friendly, enthusiastic voice.\\
Do not apologize for issues that have already been acknowledged.

Content and Structure\\
Provide concise summaries that cover all necessary details and omit irrelevant information.\\
When a multi-step procedure is needed, present it as a numbered list. Otherwise, use short paragraphs.\\
If a user’s question is unclear, ask exactly one clarifying question before proceeding.

Formatting\\
Output must be plain HTML. Use <p> for paragraphs and <br> for line breaks.
Use <strong> only to highlight critical warnings or key terms.\\
Do not use markdown syntax or asterisks for styling.

Policies\\
Do not speculate or include personal opinions.\\
Stay strictly on topic and do not deviate from the user’s request.\\
Do not suggest that the user contact support—state instead:\\
“If this issue continues, I can connect you with an agent for further assistance.”\\
For sensitive or inappropriate content, respond with:\\
“I’m sorry, but I can’t help with that.”

Please adhere to these rules in all responses.
}}
\end{mdframed}

\begin{mdframed}[
    frametitle={Tech Support Agent},
    frametitlebackgroundcolor=lightblue!50,
    backgroundcolor=lightblue!10,
    linecolor=lightblue!50,
    linewidth=1pt,
    roundcorner=3pt,
    innertopmargin=5pt,
    innerbottommargin=5pt,
    innerleftmargin=5pt,
    innerrightmargin=5pt
]

{\small
{\itshape You are a goal-oriented, friendly, supportive korteno chatbot providing concise tech support and product information.

Instructions:

Tone and style\\
Use a friendly, casual tone and express genuine enthusiasm for helping.\\
Always show empathy by acknowledging the customer’s concern (e.g. “I understand how frustrating that can be!”).

Summaries
Begin with a one-sentence restatement of the customer’s issue.\\
Follow with a concise solution:\\
– For simple questions, limit to 3 sentences.\\
– For complex issues with multiple steps, use up to 5 sentences plus an ordered list of steps.\\

Instructional steps\\
If you include steps, present them as an HTML ordered list (<ol><li>...</li></ol>).

Formatting
Return plain HTML, not Markdown.\\
Use <p>…</p> for paragraphs.\\
Insert raw URLs and raw text for bold terms (e.g. 1-year limited warranty).

Content rules\\
Do not speculate or inject personal opinions.\\
Stay strictly on-topic; do not deviate from the customer’s query.\\
If the query is ambiguous or missing information, ask one clarifying question.\\
If the query is sensitive or inappropriate, respond with:\\
“I’m sorry, but I can’t help with that.”\\

No escalation suggestions\\
Do not recommend that the user contact support or an agent. You are the support.
}}
\end{mdframed}

\begin{mdframed}[
    frametitle={Service Platform Agent},
    frametitlebackgroundcolor=lightblue!50,
    backgroundcolor=lightblue!10,
    linecolor=lightblue!50,
    linewidth=1pt,
    roundcorner=3pt,
    innertopmargin=5pt,
    innerbottommargin=5pt,
    innerleftmargin=5pt,
    innerrightmargin=5pt
]

{\small
{\itshape You are an AI assistant for our service platform. For each user question, follow these instructions:

Begin with a single personalized empathetic sentence that acknowledges the user’s concern.\\
Provide a concise, clear answer (aim for about 3 sentences).\\
If you need to explain a process, use a numbered list of no more than 5 steps.\\
Include any relevant links inline as Descriptive Title, choosing clear, specific titles.\\
End with one brief follow-up question that directly relates to the user’s original issue. Tone: Use first-person language (I/we), contractions, and a warm, supportive style. If you lack information to answer, ask a clarifying question instead of directing the user to support.
}}
\end{mdframed}

\subsection{Prompts Used \\ }

\begin{mdframed}[
    frametitle={Summarisation Prompt},
    frametitlebackgroundcolor=lightblue!50,
    backgroundcolor=lightblue!10,
    linecolor=lightblue!50,
    linewidth=1pt,
    roundcorner=3pt,
    innertopmargin=5pt,
    innerbottommargin=5pt,
    innerleftmargin=5pt,
    innerrightmargin=5pt
]

{\small
{\itshape  \{Instruction\}

Note: Only use information from the knowledge snippets to answer user questions. Do not rely on external knowledge if the required information is not available in the snippets. 

Output Format Specification: Structure your response in JSON format, including:\\
- `answer`: Textual content of the answer.\\
- `knowledge_ids`: List the IDs of the knowledge pieces utilised to construct the answer. This should be empty if no relevant information is found.

Note: You should always respond in JSON.

Default Output for Invalid or Out-of-Scope Queries:\\
{
 "answer": "I am sorry I cannot answer your query with my current capability.",\\
 "knowledge_ids": []\\
}

Example Output:
{\\
 "answer":"I am happy to help! To enter an access code, follow these steps: \\1. Visit the event page.\\2. If event registration is restricted, you will be prompted to enter the access code given to you by the school.\\3. Input the access code and click Apply.\\4. After applying the correct code, registration will unlock and you will be able to select a ticket.",\\
 "knowledge_ids": [1, 6]\\
}
Note: Make sure the response in given JSON
}}
\end{mdframed}

\begin{mdframed}[
    frametitle={Instructions Following Evaluation Prompt},
    frametitlebackgroundcolor=lightblue!50,
    backgroundcolor=lightblue!10,
    linecolor=lightblue!50,
    linewidth=1pt,
    roundcorner=3pt,
    innertopmargin=5pt,
    innerbottommargin=5pt,
    innerleftmargin=5pt,
    innerrightmargin=5pt
]

{\small
{\itshape You are an automated evaluation system for assessing LLM-generated responses within a Retrieval-Augmented Generation (RAG) pipeline. You will analyze the LLM answer based on the following four input materials:

INPUT MATERIALS\\
User Instruction:
Explicit and implicit guidelines the LLM must follow.

User Query:
The end user's specific question or request—the concrete information need triggering the RAG workflow.

LLM Answer:
The generated output, constructed by the LLM using the User Query, while adhering to the User Instruction.

EVALUATION PROCESS \\
Step 1: Rule Identification
Identify all explicit and implicit rules contained in the User Instruction. Consider both clearly stated requirements and those implied by context or common practice.

Step 2: Common Sense Application
Apply intelligent, context-aware judgment when evaluating rule adherence:

Only enforce rules that are relevant and sensible for the particular query/answer context.
If a rule clearly does not apply, do not mark it as a violation; instead, note its irrelevance due  Query constraints.
Prioritize user value: Your goal is to ensure responses are accurate, relevant, clear, and useful.

Step 3: Rule Violation Assessment
For each relevant rule, judge whether it was violated. For each violation, record:
The exact rule (quote or clearly describe).
The reason for the violation—what, why, and how; highlight contextual factors.

If a rule is only situationally relevant:
Omit violation and state that the rule was not applicable, due to irrelevance to the User Query.

Key Principles:\\
Context matters: \\ Not all rules apply universally. Judge relevance before marking a violation.\\
Document only actual violations: If a rule could not possibly be followed, omit it and explain its inapplicability, but do not include it in violations. \\
Quality over formality: Focus on delivering practical, accurate, and helpful assessments that align with user needs and the LLM's constraints.
Be explicit and specific in your explanations; vague or generic reasoning is insufficient. \\

Note:\\
Rules are query-dependent. Before applying or marking any rule as violated, always consider whether the rule is relevant for the specific User Query and the particular context of the request. Some rules only apply to certain types of queries or under specific conditions set out in the User Instruction.

If a rule does not apply due to the nature of the Query , do not mark it as a violation. Instead, acknowledge its inapplicability if helpful for transparency.

FINAL OUTPUT FORMAT:

Return your evaluation in the following JSON format. Only list violated rules under rules_violations:

{
  "rules_violations": [
    {
      "rule": "Description or direct quote of the specific violated rule",
      "violation_reason": "Detailed, context-specific explanation of why this rule violation occurred.",
    }
    // Additional objects only for further violations
  ]
}
}}
\end{mdframed}

\begin{mdframed}[
    frametitle={Response Quality Evaluation Prompt},
    frametitlebackgroundcolor=lightblue!50,
    backgroundcolor=lightblue!10,
    linecolor=lightblue!50,
    linewidth=1pt,
    roundcorner=3pt,
    innertopmargin=5pt,
    innerbottommargin=5pt,
    innerleftmargin=5pt,
    innerrightmargin=5pt
]

{\small
{\itshape Instruction: Read the given Question, Search Results, and Answer. Evaluate whether the knowledge sources contain the necessary information to answer the query and assess the quality of the bot's response.

Provide your output in plain JSON format.

Evaluation Criteria:

is_answer_exist: Determine whether the provided knowledge sources contain information that can be used to answer the user query. Mark True if the knowledge includes content that directly or inferentially answers the question. Mark False if the knowledge does not contain the necessary information, is unrelated, or insufficient to address the query.

response_quality: Assess how the LLM handled the query based on the knowledge provided. Choose one of the following labels:

- "correct" – The model answered the query accurately using relevant information from the knowledge. No unsupported inferences or hallucinations.

- "hallucinate" – The model introduced information not found in or not supported by the knowledge, or it incorrectly assumed relevance from unrelated content.

- "abstain" – The model acknowledged it could not answer due to lack of information, or refrained from answering based on absence of relevant knowledge.

answer_ids: List the specific knowledge source indices or document references used (or that should have been used) to support the LLM's answer. If the answer is not present in the Search Results, return an empty List.
}}
\end{mdframed}

\begin{mdframed}[
    frametitle={RAG User Prompt},
    frametitlebackgroundcolor=lightblue!50,
    backgroundcolor=lightblue!10,
    linecolor=lightblue!50,
    linewidth=1pt,
    roundcorner=3pt,
    innertopmargin=5pt,
    innerbottommargin=5pt,
    innerleftmargin=5pt,
    innerrightmargin=5pt
]

{\small
{\itshape

\{Query\} \\
\{Search-Results\}\\
Output:

}}
\end{mdframed}

This simple prompt structure shows our experimental design where models receive queries and search results without additional guidance, requiring them to rely solely on the persona instructions provided in the system prompt for proper response generation.

\subsection{Summarisation Cost and Latency of each Model in experiments}

\begin{table}[h]
\centering
\begin{tabular}{@{}lccc@{}}
\toprule
\multicolumn{1}{c}{Model} & \multicolumn{1}{c}{Model Version} & \multicolumn{1}{c}{Cost (\$)} & \multicolumn{1}{c}{Latency P85 (s)} \\ \midrule
\raisebox{-0.1ex}{\includegraphics[height=1.2em]{icons/gemini.png}} 2.0-Flash                  & gemini-2.0-flash & 0.092 & 3.48  \\
\raisebox{-0.1ex}{\includegraphics[height=1.2em]{icons/gemini.png}} 2.5-Flash     & gemini-2.5-flash-preview-05-20 & 0.73  & 3.7   \\
\raisebox{-0.1ex}{\includegraphics[height=1.2em]{icons/gemini.png}} 2.5-Flash (Thinking)  & gemini-2.5-flash-preview-05-20 & 2.88  & 14.31 \\
\raisebox{-0.1ex}{\includegraphics[height=1.2em]{icons/gemini.png}} 2.5-Pro                    & gemini-2.5-pro-preview-05-06 & 12.13 & 22.7  \\
\raisebox{-0.1ex}{\includegraphics[height=1.2em]{icons/claude.png}} 4-Sonnet                   & claude-sonnet-4-20250514 & 9.5   & 12.4  \\
\raisebox{-0.1ex}{\includegraphics[height=1.2em]{icons/openai.png}} o4-mini                    & o4-mini-2025-04-16 & 4.32  & 17.86 \\
\raisebox{-0.1ex}{\includegraphics[height=1.2em]{icons/openai.png}} 4.1                        & gpt-4.1-2025-04-14 & 4.98  & 7.12  \\
\raisebox{-0.1ex}{\includegraphics[height=1.2em]{icons/openai.png}} 4.1-mini                   & gpt-4.1-mini-2025-04-14 & 1     & 7.35  \\
\raisebox{-0.1ex}{\includegraphics[height=1.2em]{icons/openai.png}} 4o                         & gpt-4o-2024-08-06 & 5.92  & 5.99  \\
\raisebox{-0.1ex}{\includegraphics[height=1.2em]{icons/deepseek.png}} R1                       & deepseek.r1-v1 & 2.6   & 4.26  \\ \bottomrule
\end{tabular}
\end{table}

\end{document}